%% file: main.tex
\documentclass[twocolumn,letterpaper]{article}
\usepackage{fullpage}
\usepackage{amsmath,amsfonts}
\usepackage{algorithmic}
\usepackage{algorithm}
\usepackage{array}
\usepackage[caption=false,font=normalsize,labelfont=sf,textfont=sf]{subfig}
\usepackage{textcomp}
\usepackage{stfloats}
\usepackage{url}
\usepackage{verbatim}
\usepackage{graphicx}
\usepackage{xcolor}
\usepackage{cite}
\usepackage[colorlinks,bookmarksopen,bookmarksnumbered,citecolor=red,urlcolor=red]{hyperref}

\newcommand{\Real}{\mathbb{R}}
\newcommand{\e}{\mathbf{e}}
\newcommand{\Hf}{\mathcal{H}}
\newcommand{\SpX}{\mathfrak{X}}
\newcommand{\SpY}{\mathfrak{Y}}
\newcommand{\se}{\mathfrak{se}}
\newcommand{\SE}{\mathsf{SE}}
\newcommand{\La}{\mathsf{L}}
\newcommand{\T}{\mathsf{T}}
\newcommand{\ODW}{Ostwald--de\,Waele}

\begin{document}

\title{Linear Motility Maps in Nonlinear Viscous Fluids}

\author{
Yishun Zhou and Shai Revzen
\thanks{Y. Zhou is in the Department of Robotics, University of Michigan, Ann Arbor, MI, USA.}
\thanks{S. Revzen is in the Departments of Electrical Engineering and Computer Science, and Ecology and Evolutionary Biology, University of Michigan, Ann Arbor, MI, USA.}
}

\markboth{}{}

\maketitle

\begin{abstract}
    Systems moving in low Reynolds number fluid regimes are known to be governed by a ``motility map'' which linearly relates their shape change rates to they body frame velocity moving through the fluid.
    A consequence of this is ``Purcell's Scallop Theorem'' -- a locomotion system that undergoes shape changes that follow the same path forward and backward in time (reciprocal body deformations) cannot achieve net displacement, regardless of pacing of those changes.
    We show that linear-in-velocity motility maps extend to any power law viscosity (a.k.a. Ostwald--de Waele fluid), and therefore to many biological fluids in intermediate shear ranges. 
    We also show that the linear-in-velocity property can be violated in Carreau-Yasuda fluids to produce net motion using an ``inchworm'' model consisting of two unequal masses with unequal drag coefficients performing reciprocal motions.
    Interestingly, the direction of motion can be switched by changing speeds.
    Our results show that the linear motility map of geometric mechaincs can be used to analyze and design locomotion in power-law fluids, and that some nonlinear drag relationships such as Carreau-Yasuda can be exploited to generate net locomotion in seeming violation of the ``scallop theorem''.
\end{abstract}

\input{introduction}

\input{homogeneous}
\input{locomotion_model}

\input{discussion}

\section*{Acknowledgments}
The authors would like to acknowledge Eleni Gourgou and Ross Hatton for the initial motivation to explore this avenue of investigation.

\bibliographystyle{IEEEtran}
\bibliography{reference}

\end{document}

%% file: introduction.tex
\section{Introduction}
The ``Purcell Scallop Theorem'' states that at the limit of zero Reynolds number, a swimmer cannot achieve net displacement through a cycle of reciprocal motion in a Newtonian fluid\cite{purcell1977}. 
In these low Reynolds number regimes the ``local connection'' \cite{kelly1995}, more recently renamed the ``motility map'' \cite{bass2022characterizing}, accurately describes the relationship between body shape change and body motion through space. 
It is a function of the form $A(r,\dot r) \mapsto v_b$ mapping shape $r$ and shape change rates $\dot r$ to body frame velocities $v_b$.
Furthermore in the conventional linear viscosity case it is linear in $\dot r$ and therefore often written as $v_b=A(r)\cdot \dot r$.

Under the assumption that such a linear motility map governs motion, the classical ``Purcell Scallop Theorem'' is: if the locomoting body's shape performs a reciprocal motion, the result is no net motion. 
Our detailed proof of a (generalized) Scallop Theorem is in sec. \ref{sec:pfScal} below.
In this classical form the Scallop Theorem follows from the linearity of the motility map, which is itself proven for linear viscosity (e.g. \cite{kelly1995}). 
Therefore it may seem that nonlinear friction laws, such as Coulomb friction and power-law drag could break the Scallop Theorem and allow net displacement with reciprocal gaits. 
The core of this paper focuses on considernig whether such Scallop Theorem violations occur, and under what regimes.

There are many friction and drag models used for different kinds of fluids and contacts.
Because of the interrelationship of these concepts we will use the terms ``drag'', ``viscosity'', and ``friction'' interchangeably -- they all refer to reaction forces generated by a body moving while in contact with other matter in its environment.

Of the friction laws, the most familiar is Coulumb friction \cite{coulomb1785}, a.k.a. ``dry friction'', which is considered a good model for mechanical contacts between solids. 
Coulomb friction is a non-viscous friction.
It is independent of velocity, and has a static threshold force preventing motion.
When motion while in contact exists, it predicts a dynamic force that depends only on the contact velocity's direction but not on its magnitude.
For fluids, linear drag or linear viscosity are the textbook model of Newtonian fluid reactions to a body moved through it at ``low enough'' speeds.
Power-law drag appears in some fluids.
It is also called ``Ostwald-deWaele'' viscosity \cite{macosko1994}.
Here the drag force follows a power law with respect to the magnitude of the velocity, typically used to capture shear thinning or thickening phenomena.
More complex fluids are better approximated by different power law exponents at low and high shear rates. 
For example, the Carreau-Yasuda model \cite{macosko1994} captures the time-invariant viscosity behavior of many biological fluids where the fluid behaves like a Newtonian fluid at low and high shear rates but follows a power law in the intermediate shear rate regime. 
Finally, another class of drag models is ``viscoelastic drag''.
This class of models includes terms reacting to velocity, but also time-dependent terms.
We consider viscoelastic drag outside the scope of the current paper.

The nonlinear viscosity models mentioned are often cited as good models for many biological fluids.  
For example, the power-law model approximates the visocity of the cytoplasm of human neutrophils \cite{tsai1993}. 
Some biological fluids can be effectively modeled as power-law fluids when considering only a limited range of shear rates \cite{lynch2022}.
The more generalized Carreau-Yasuda model, is widely used to describe shear-thinning biological fluids including blood \cite{mekheimer2018carreau,Gataa2024} and mucus \cite{Bartlett2023} because it recovers a power-law regime at intermediate shear rates while also capturing low-and high-shear behaviors. 
These nonlinear viscosity models help describe how microorganisms and microswimmers interact with their environment and can provide insights into their locomotion. 
Several studies of the locomotion of microswimmers in nonlinear viscosity fluids showed that this different regime, compared to the linear drag Newtonian fluid regime, has varied effects: it can have no effect on locomotion speed \cite{gagnon2014}, change the magnitude and direction of movement \cite{Qin2023}, or increase and decrease locomotion velocity \cite{Qin2021, Gomez2016}. 

Here we present the results of our work, which includes theoretical proofs that the linear-in-shape-velocity motility map extends outside the familiar linear drag regime to \ODW\ fluids.
Because the motility map applies, the Scallop Theorem too applies to power law viscosity fluids, and (in a limiting sense) even to Coulomb friction.
By exploring the behavior of a 1-DoF ``inchworm'' undergoing reciprocal shape-change, we also examine the effects of inertia to drag ratios (``Reynolds Number'') and the type of fluid viscosity model used: linear, power-law, and Carreau-Yasuda.
We found the expected Scallop Theorem results for linear and power-law viscosities, and found that in the Carreau-Yasuda fluid a change of speed can produce a change of direction of motion.

%% file: homogeneous.tex
\section{Homogeneous Functions and Viscosity}\label{sec:homogeneous}
In this section of the paper we will prove some mathematical results about the properties that arise from certain classes of functions that could be used as viscosity laws.

Since we wish to examine how broad a class of viscosity laws the Scallop Theorem would apply to, let us consider a (seemingly) general scaling property of the viscosity -- that there exists a scaling law function $g(s)$ such that for any scale $s\in\Real$ and body part moving with velocity $\dot x$, the friction force $f()$ scales with $\dot x$ by
\begin{align}
	f(s \dot x) = g(s) f(\dot x).
\end{align}

Treating this as a function equation (i.e. an equation whose unknowns are functions), we can obtain some additional properties: $g(1)=1$, and $f(s) = g(s) f(1)$, or
\begin{align}
	g(a) g(b) f(1) = g(a) f(b 1) & = f(a b 1) = g(a b) f(1)  ,\nonumber\\
	g(a b) &= g(a) g(b),
\end{align}
i.e. if $f(1) \neq 0$ then the function $g$ is completely multiplicative.

The question of completely multiplicative functions on the reals is a classical one, and has been solved.
Substituting $g(x) =: \e^{h(\ln(x))}$, $a := \e^v$ and $b := \e^u$ we get
\begin{align}
	g(a b) &= g(a) g(b)\nonumber\\
	\e^{h(v+u)} &= \e^{h(v)}\e^{h(u)} = \e^{h(v)+h(u)}, \nonumber\\
	h(v+u)&=h(v)+h(u).
\end{align}
Solving for which functions $h$ satisfy this equation is called ``Cauchy's Function Equation''.
It is known that the solutions are either linear or nowhere continuous.
Since a nowhere continuous function is non-physical, only the linear solutions are relevant as viscosity models.
They are of the form $h(x) := \gamma x$, and lead to a power law viscosity, i.e. to $f(s \dot x) = s^\gamma f(\dot x)$ -- an \ODW\ fluid.
We conclude that the only possible viscosity models that have a ``general scaling law'' as defined above are the \ODW\ models.

\subsection{Properties of Homogeneous Functions}\label{sec:propHf}

Consider a pair of real vector spaces $\SpX$ and $\SpY$, and some function $f: \SpX \to \SpY$. 
We define the set $\Hf^\gamma$ of ``(real) homogeneous functions of order $\gamma$'' as the functions which satisfy $f(sx) = s^\gamma f(x)$ for all $s\in\Real$ and $x \in \SpX$.
The spaces $\SpX$ and $\SpY$ should be clear from the context.
If the order of homogeneity is not noted, we mean this order to be 1.

We shall note a few important properties of $\Hf^\gamma$ functions.

\paragraph{Composition Rule}\label{sec:hcomp}
If $f\in \Hf^\gamma$ and $g \in \Hf^\beta$, then $f \circ g \in \Hf^{\gamma\beta}$ wherever it is defined.

\paragraph{$\Hf^\gamma$ is a linear function space}
If let $f,g \in \Hf^\gamma$ and $a,b \in \Real$, then $a f + b g\in  \Hf^\gamma$.

\paragraph{Linear Functions are Homogeneous}
All linear functions are in $\Hf^1$.

\paragraph{Homogeneous Constraints have Homogeneous Solutions}
Consider a constraint equation of the form $f(x+y)=0$ for $f\in\Hf^\gamma$.
If the equation has a solution $(x_0,y_0)$, then it is also solved by $(\alpha x_0, \alpha y_0)$ for all $\alpha \in \Real$ because $f(\alpha x_0+ \alpha y_0) = f(\alpha\,(x+y)) = \alpha^\gamma f(x+y) = 0$.
It follows that for any implicit solution function $\tilde y(\cdot)$ such that $f(x+\tilde y(x))=0$, $\tilde y \in \Hf^1$ i.e. the implicit functions defined by homogeneous constraints are homogeneous of order 1.

\subsection{Application to Friction Dominated Motion}

When friction forces dominate inertia, the motion we consider is quasi-static, i.e. we assume that the total forces and torques on the body are zero.
This includes the well known case of low Reynolds number swimming, and also the less well-known cases of slithering \cite{Hu2009} and multi-legged locomotion \cite{zhao2022} thought to be governed by dynamic Coulomb friction.

For simplicity we will further assume a stationary environment, which the moving body interacts with through a finite number of ``contacts''.
These could represent actual contact points, blade elements in a fluid dynamics model, or any other components which produce reaction forces from the environment by means of their velocity relative to the material in the environment which is not moving in the world frame.

Assuming the positions of the contacts in the body frame are given by $\{r_k\}_{k=1}^{N_c}$ and that the body frame is given by the rotation $R(t)$ and translation $p_0(t)$, the contacts are
\begin{align}
	p_k(t) := R(t) r_k(t) + p_0(t).
\end{align}
From here on we elide the time dependence $\square(t)$ for clarity of notation.
The body translation velocity $v_b := \dot p_0$.
The body attitude velocity is $\dot R = R [\omega]_\times$ where $[a]_\times$ is the ``cross product'' matrix such that $[a]_\times b = a \times b$, in notation similar to that of \cite{lynch2017}.
This leads to contact point velocities 
\begin{align}
	\dot p_k = R \dot r_k + R [\omega]_\times r_k + v_b
	= R \dot r_k - R [r_k]_\times \omega + v_b.
\end{align}
Notably, these are functions of $r$, $\dot r$, $v_b$, and $\omega$ which are linear in $\dot r$, $v_b$, and $\omega$, and therefore $\Hf^1$.

Assume contacts are governed by the same viscosity law $f(\cdot) \in \Hf^\gamma$ which acts in a position dependent manner, e.g. via a drag matrix $\mu_k(r)$, then
\begin{align}\label{eqn:forces}
	F & = \sum_k \mu_k(r) f( \dot p_k ) \nonumber\\
	M & = \sum_k [r_k]_\times \mu_k(r) f( \dot p_k ),
\end{align}
which is therefore homogeneous of order $1 \cdot \gamma \cdot 1 = \gamma$. 

At the limit of zero Reynolds number, the net force and torque on the body must be zero, i.e. $F = 0$ and $M=0$. Given the observations from \nameref{sec:propHf} regarding constraints, it follows that holding $r$ constant, any implicit mapping from $\dot r$ to $(\omega, v_b)$ would be $\Hf^1$.
The motility map is the totality of such implicit mappings over all values of $r$: $A(r,\dot r) := (\omega, v_b)$.
Thus we conclude that \textbf{for any \ODW\ fluid, the motility map $A(r,\dot r)$ is homogeneous  in $\dot r$}.

\section{$A(r,\cdot) \in \Hf^1$ implies the Scallop Theorem}\label{sec:pfScal}

In this section we provide a proof of the Scallop Theorem for the case $A(r,\cdot) \in \Hf^1$, which generalizes the classical case where $A(r,\cdot)$ is linear.

\paragraph{Definition of ``(Smooth) Reciprocal Motion''} a reciprocal motion is any smooth shape change that can be realized as a pair of functions $r_0$ and $\tau$. 
The $r_0$ function is a ``canonical reciprocal motion'' mapping $[0,2]\subset \Real$ into shape, and satisfying $r_0(1+t)=r_0(1-t)$.
The $\tau$ function is the ``rate'' of the motion: $\tau:[0,T] \to [0,2]$, $\tau(0)=0$, $\tau(T)=2$. 
The reciprocal motion is the body shape change given by the composition $r(t) := r_0(\tau(t))$.

\paragraph{The Scallop Theorem for $\Hf^1$ Motility Maps} For any locomotion system whose motion is governed by a motility map $v_b=A(r,\dot r)$ which is $\Hf^1$ in $\dot r$, and any smooth reciprocal motion of that system's body, the result is no net motion. 

\paragraph*{Proof} Take the body frame of the moving system to be a time varying $g \in \SE(3)$, and $\La_g:\se(3) \to \T_g$ to be the left action of $g$ mapping the body velocities as Lie Algebra elements to the tangent space of $g$. (computational note: in the notation of \cite{lynch2017} $\La_g(v) := g \cdot [v]$.)
The net motion produced by a shape change trajectory $r: [0,T] \to \SpX$ is given by solving
\begin{align}
	\dot g = \La_g A(r,\dot r); && g(0)=\text{Id}. \label{eqn:fwd}
\end{align}
For a given shape-change $r(t)$, the motility map is bounded, and therefore the RHS of equation \ref{eqn:fwd} is Lipschitz and equation \ref{eqn:fwd} has a unique solution that extends for all of $[0,T]$.
 
Now consider a rate change, where instead of $r(t)$ we use $r(\tau)$ for some differentiable $\tau(t)$.
From the chain rule $\frac{d}{dt}r(\tau) = \dot\tau \dot r(\tau)$.
From $A(r,\cdot)\in\Hf^1$ we have $A(r,\dot\tau\dot r)=\dot\tau A(r,\dot r)$.
By inspection, the time rescaled body motion in space $g(\tau)$, solves this new equation, since 
\begin{align}
	\dot \tau \dot g = \dot \tau \La_g A(r,\dot r) = \La_g A(r, \dot \tau \dot r),
\end{align}
and this is the only solution, from uniqueness.

Because the time parameterization $\tau$ can be chosen arbitrarily, we need only prove the Scallop Theorem for canonical reciprocal motions $r$, i.e. $r$ is defined on $[0,2]$ and satisfying $r(1+t)=r(1-t)$ for $t\in[0,1]$.

This $r(t)$ motion has some smoothly time dependant body velocity $v_b(t)$ for $t\in [0,1]$, and $\dot g = \La_g v_b$.
Consider $h(t) := g(1-t)$.
We have 
\begin{align}
  \dot h & = \dot g(1-t) \cdot (-1) \\
  & = - \La_{g(1-t)} v_b(1-t) \\
  & = - \La_{h(t)} v_b(1-t).
\end{align}
Consider now that 
\begin{align}
    v_b(1+t) & = A(r(1+t),\dot r(1+t)) \\
    & = A(r(1-t),- \dot r(1-t)) \\
    & = -A(r(1-t),\dot r(1-t)) = -v_b(1-t).
\end{align}
This implies that $\dot h =  \La_{h(t)} v_b(1+t)$, i.e. that $h$ is the solution for the return stroke of the motion, and $g(1+t)=g(1-t)$, proving that $g(2)= g(0)$ and no net motion could have been produced. \textbf{Q.E.D.}

In this section we demonstrated two key results:

\paragraph*{Power Law Friction gives Motility Maps that are Homogeneous} We have shown that a moving body that propels itself by reaction forces that are $\Hf^\gamma$ homogeneous of order $\gamma$ in the contact sliding speeds will have a motility map $A(r,\dot r)$ which is $\Hf^1$ homogeneous in $\dot r$.
This implies that all \ODW\ fluids produce homogeneous motility maps when the Reynolds number approaches 0.

\paragraph*{Reciprocal Motion Governed by a Homogeneous Motility Map Produces No Motion} We have proven a generalization of Purcell's Scallop Theorem to all motions in systems with motility maps that are homogeneous in the rate of shape change. 
This includes \ODW\ fluids and non-fluid systems governed by power-law friction.

\section{Coulomb Friction is a limiting case of Power Law Friction}

One persistent puzzle of motility maps has been the fact that these maps, historically arising from linear viscous systems, seem to be effective for modeling multi-legged systems subject to Coulomb friction \cite{zhao2022, wu2019, wu2025}.
Our results here may throw some additional light on this question.

Consider the Lagrangian formulation of a locomotion system subject to Coulomb friction
\begin{align}
	\frac{d}{dt} \left(\frac{\partial\mathcal{L}}{\partial q}\right) - \left(\frac{\partial\mathcal{L}}{\partial \dot q}\right) = F,\label{eqn:lagPDE}
\end{align}
where the non-conservative dissipation forces are given by $F$.
The Coulomb friction contacts split into static contacts which do not dissipate energy and therefore would not be included in $F$, and dissipative dynamic friction contacts whose forces are of the form $f(p_k) = f(p_k/\|p_k\|)$.
Typically, $F$ is a smooth function of these $f(p_k)$ contact forces.

If we replace $f(p_k)$ with an $\Hf^\gamma$ relaxation $f_\gamma(p_k) := f(p_k/\|p_k\|) \|p_k\|^\gamma$, the resulting disspation force $F_\gamma$ would have the property that $F = \lim_{\gamma \to 0} F_\gamma$.
To the extent the solutions of \ref{eqn:lagPDE} modified by $F_\gamma$ depend conntinuously on $\gamma$, the resulting system dynamics with Coulomb friction are the limit of system dynamics with power-law friction.
If the power-law relaxation solutions follow a motility map $A_\gamma(r,\dot r)$, a limiting motility map at $\gamma \to 0$ may exist and be an accurate model of the original Coulomb friction system.

%% file: locomotion_model.tex
\section{A Minimal ``Scallop'' Model}\label{sec:locomotion_model}

To examine the appearance of Scallop Theorem-like results in other fluid drag models, we needed to produce a dynamical model which could be taken to the zero Reynolds number limit.
This means producing a family of models which have progressively smaller inertial forces relative to viscous forces, yet execute (inasmuch as possible, at least) geometrically similar motions. 
We realized the effect of decreasing inertial forces and thereby Reynolds number by decreasing the mass of the system while keeping the shape change profile and the drag law fixed.

\subsection{The Inchworm Model}
\begin{figure}[h]
    \centering
    \includegraphics[width=0.8\linewidth]{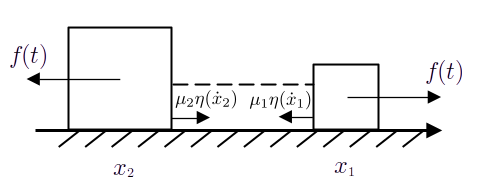}
    \caption{Inchworm model schematic. %
	Two masses $m_1$ and $m_2$ are positioned at $x_1$ and $x_2$ respectively. %
	An actuation force $f(t)$ acts on the two masses in opposite directions. %
	The drag forces act in the opposite direction of the velocity of each mass with drag coefficients $\mu_1$ and $\mu_2$. %
	The function $\eta$ represents the friction model we choose.}
    \label{fig:inchworm}
\end{figure}
As our minimalistic model, we consider the one-dimensional dynamics of two masses with disparate magnitudes and drag coefficients.
The dynamics are driven by an actuator force $f$ and resisted by drag
\begin{align}
	m_1 \ddot x_1 = f - \mu_1 \eta(\dot x_1) && m_2 \ddot x_2 = -f - \mu_2 \eta(\dot x_2). \label{eqn:dynamics}
\end{align}

For the classical viscous drag case, $\eta(v)= v$, which is the special $\beta=1$ case of an \ODW\ fluid
\begin{align}
	\eta(v) = v |v|^{\beta-1}. \label{eqn:odw}
\end{align}
For a Carreau-Yasuda fluid model the term is more complex
\begin{align}
	\eta(v) = v \left( \eta_\infty + (\eta_0 - \eta_\infty)(1+(\lambda |v|)^a)^{\frac{n-1}{a}}\right). \label{eqn:carr}
\end{align}

\begin{figure}
	\centering
    \includegraphics[width=0.8\linewidth]{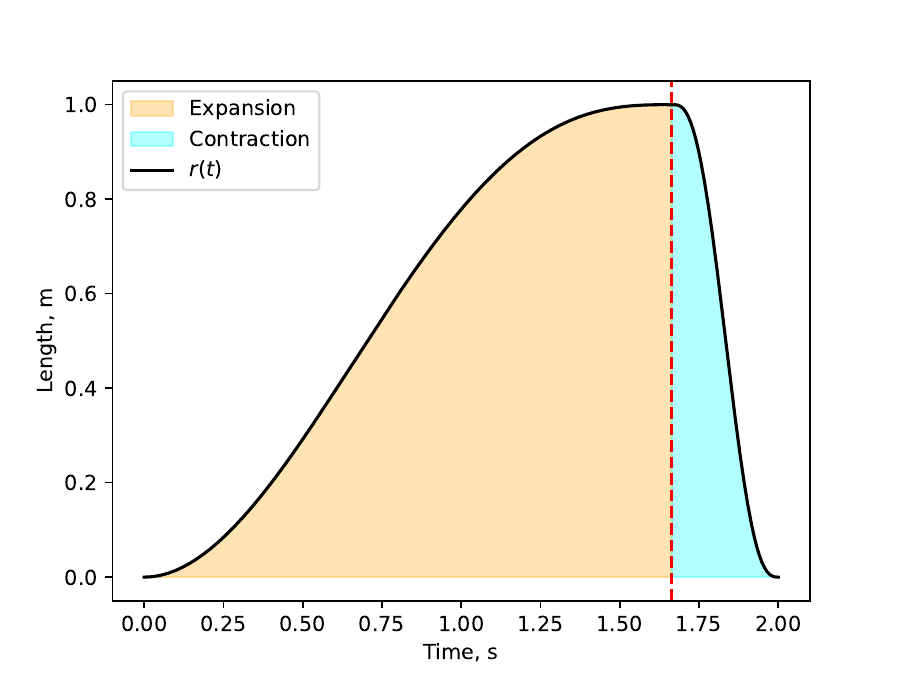}
    \caption{Inchworm shape change profile over time. %
	  We pieced together two hermite splines of the form of equation \ref{eqn:shape}. }
    \label{fig:shape}
\end{figure}
We prescribed a shape change profile $r^*(t) = x_1(t)-x_2(t)$ by piecing together two 5th-order hermite polynomials, while ensuring they were joined smoothly to second order (i.e. the resulting cycles are $C^2$ smooth). 
The distance $r(t)$ between the two masses expanded from length 0 to length 1, and then contracted back to length 0 in a total of 2 time units, where 
\begin{align}
	r^*(t) = \sum_{i = 0}^{n} a_i t^i. \label{eqn:shape}
\end{align}
To achieve this shape change at all time $t$, we computed the necessary actuator force $f$ at each time step via plant inversion control 
\begin{align}
	f = \frac{m_1m_2 \ddot r + m_2 \mu_1 \eta(\dot{x}_1) - m_1 \mu_2 \eta(\dot{x}_2)}{m1+m2},
\end{align}
and set the initial shape change $r(t_0) = r^*(t_0)$ and $\dot{r}(t_0) = \dot{r}^*(t_0)$. 

\subsubsection{The solution to \ref{eqn:dynamics_com} is periodic}
We took the center of mass position $x := (m_1 x_1 + m_2 x_2)/M$, where $M = (m_1+ m_2)$ and rewrote the dynamics \ref{eqn:dynamics} as
\begin{align}
	M\ddot{x} = - \mu_1 \eta(\dot{x} + \frac{m_2}{M} \dot{r}) - \mu_2 \eta(\dot{x} - \frac{m_1}{M} \dot{r}). \label{eqn:dynamics_com}
\end{align}

\paragraph{$\dot{x}$ is bounded}
If $r(t)$ was prescribed as \ref{eqn:shape}, then there exists $\dot{r}_{max}$ and $\dot{r}_{min}$. %
Then $\dot{x}$ is bounded bewteen $[\max(-\frac{m_2}{M}\dot{r}_{min}, \frac{m_1}{M}\dot{r}_{max}), \min(-\frac{m_2}{M}\dot{r}_{max}, \frac{m_1}{M}\dot{r}_{min})]$.

\paragraph{Solution $\dot{x}(t)$ exists}
Since the right hand side of equation \ref{eqn:dynamics_com} is continuous, a solution exists for any initial condition $\dot{x}(t_0)$ by the Peano existence theorem \cite{peano1886}.

\paragraph{Solution $\dot{x}(t)$ is unique}
From \cite[Ch. 1, Thm. 2, Rem. 11.]{Filippov}, the RHS of equation \ref{eqn:dynamics_com} being monotone non-increasing in $\dot{x}$ guarantees the uniqueness of the solution forward in time. 
Both power law drag in \ref{eqn:odw} with $\beta >=0$ and the Carreau-Yasuda drag model in \ref{eqn:carr} with shear thinning ($n<1$) are monotone non-decreasing in $\dot{x}$, so the solution is unique.

According to \cite[Thm. 1]{Massera1950}, there exists a periodic solution $\dot{x}(t)$ with the period of $r(t)$. 
In fact, the periodic solution for the system appears to be a stable limit cycle according to simulation. 
At $\beta = 1$, it is a special case that the net displacement over a cycle at steady state is always zero. 
When $\beta = 1$, \ref{eqn:dynamics_com} becomes
\begin{align}
	\ddot{x} + \frac{\mu_1 +\mu_2}{M} \dot{x}  = \frac{\mu_2 m_1 - \mu_1 m_2 }{M^2} \dot{r}(t) \label{eqn:linear_com}.
\end{align}
Taking integral of \ref{eqn:linear_com} over one cycle from $0$ to $T$ gives
\begin{align}
	\frac{\mu_1 +\mu_2}{M} \int_0^T \dot{x} dt = \frac{\mu_2 m_1 - \mu_1 m_2 }{(M)^2} \int_0^T \dot{r}dt - \int_0^T \ddot{x} dt 
\end{align}
where $\int_0^T \dot{r}dt = r(T) - r(0) = 0$ and $\int_0^T \ddot{x} dt = \dot{x}(T) - \dot{x}(0) = 0$ since the system is periodic. 
Therefore $\int_0^T \dot{x} dt = 0$, which means the net displacement over a cycle at steady state is always zero. 

We used numerical root finding to find the initial condition $\dot{x}(t_0)$ that leads to a periodic solution and ran each simulation for 10 cycles. 
While there were numerical errors, relative difference in net displacement per cycle eventually reached values below $10^{-4}$ within this interval of 10 cycles which we consider to be periodic to within numerical tolerances.

\begin{figure*}[!htbp]  
	\centering
	\subfloat[]{%
		\includegraphics[width=0.48\textwidth]{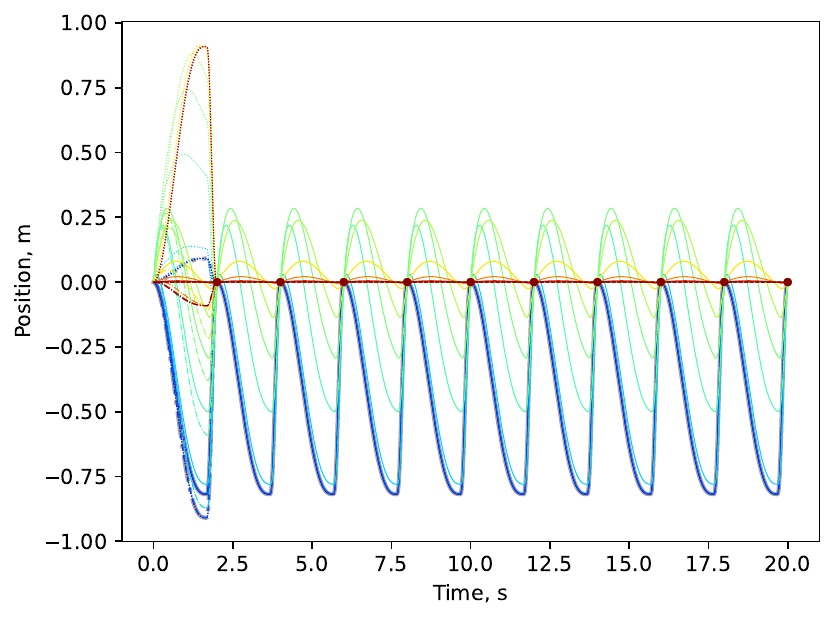}
		\label{fig:traj_n=1}}
	\hfill
	\subfloat[]{%
		\includegraphics[width=0.48\textwidth]{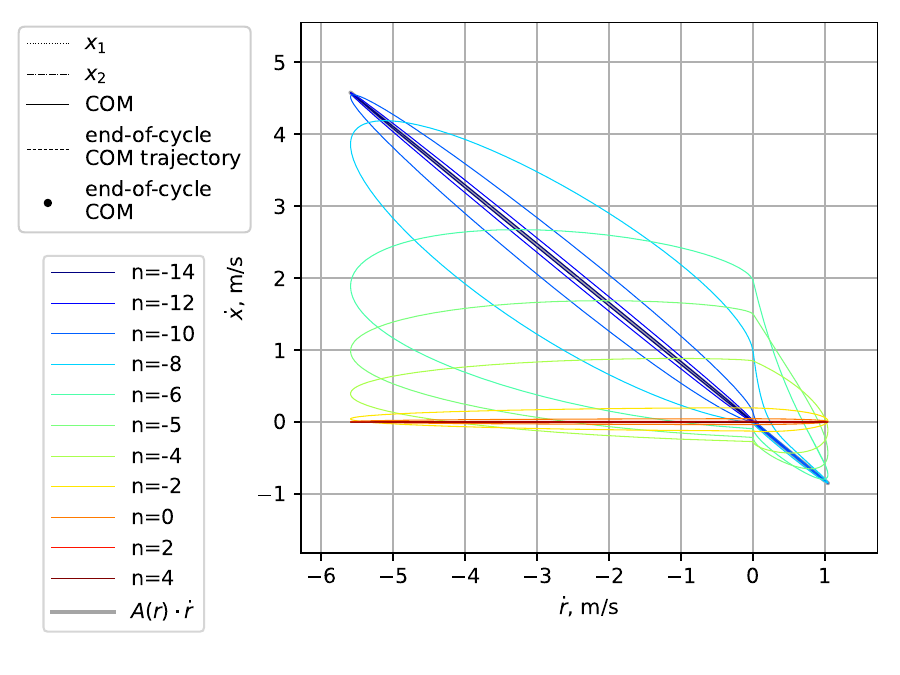}
		\label{fig:conn_n=1}}
	\caption{Inchworm dynamics with linear viscous friction at different Reynolds numbers with different masses, where $m_2 = 10m_1$ and $\mu_2 = \frac{\mu_1}{10}$. %
	(a) The positions of the two masses $x_1$ and $x_2$ versus time for different masses $m_1$, where $m_1 = 2^n$ kg. %
	The positions $x_1$ (dotted line), $x_2$ (dashed line), and the center of mass (COM) position (solid line) are shown. %
	The two masses oscillate causing the COM to move, however the net displacement over a cycle at steady state is always zero. %
	(b) COM velocity $\dot{x}$ versus shape change velocity $\dot{r}$ at different Reynolds numbers $R_e$ and at $R_e = 0$ (grey). %
	As $R_e$ decreases, the $\dot{x}$ versus $\dot{r}$ relationship gets closer to the motility map $\dot{x} = A(r) \dot{r}$, where $A = \frac{\mu_2 m_1 - \mu_1 m_2}{(\mu_1+\mu_2)(m_1+m_2)}$ is independent of $r$.}
	\label{fig:n=1}
\end{figure*}

\begin{figure*}[!htbp]  
	\centering
	\subfloat[]{%
		\includegraphics[width=0.48\textwidth]{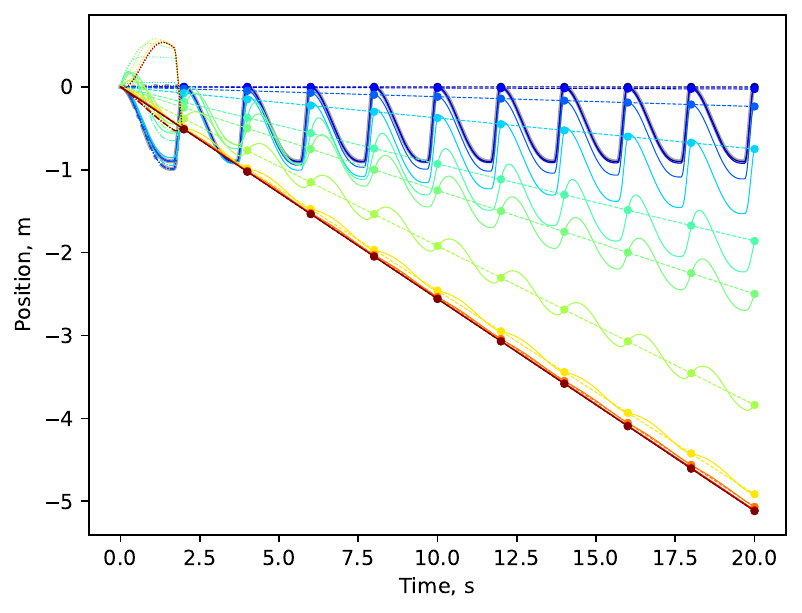}
		\label{fig:traj_n=1}}
	\hfill
	\subfloat[]{%
		\includegraphics[width=0.48\textwidth]{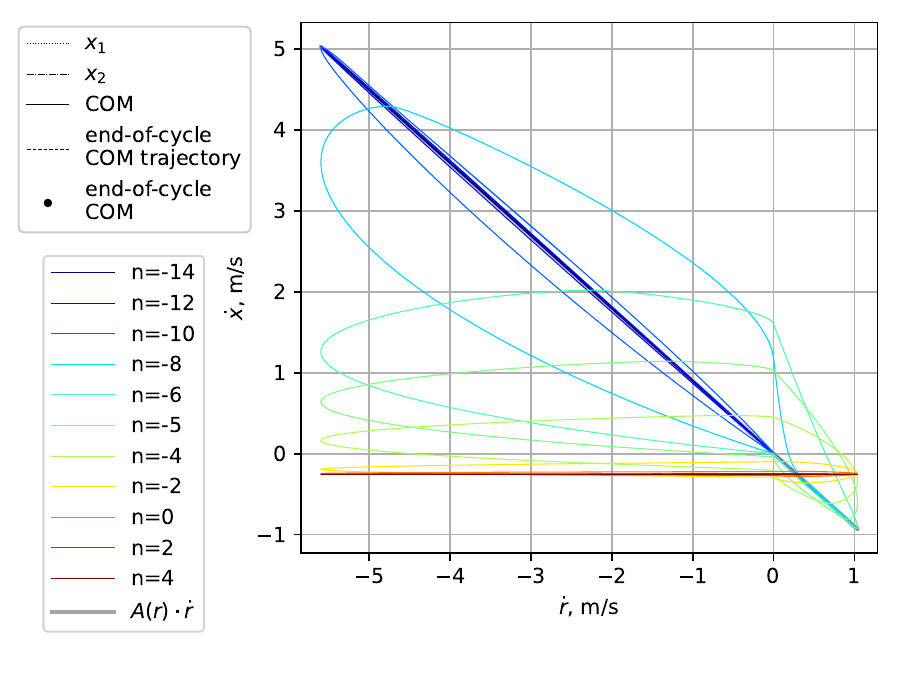}
		\label{fig:conn_n=1}}
	\caption{Inchworm dynamics with power law 1/2. %
	We simulated multiple Reynolds numbers and masses, holding $m_2 = 10m_1$ and $\mu_2 = \frac{\mu_1}{10}$. %
	(a) The positions of the two masses $x_1$ and $x_2$ versus time for different masses $m_1$, where $m_1 = 2^n$ kg. %
	The positions $x_1$ (dotted line), $x_2$ (dashed line), and the center of mass (COM) position (solid line) are shown. %
	The two masses oscillate causing the COM to move. %
	The net displacement over a cycle for small masses, i.e. Reynolds numbers $R_e\to 0$, goes to zero as expected. %
	When the masses are large, a net motion is produced. %
	(b) COM velocity $\dot{x}$ versus shape change velocity $\dot{r}$ at different Reynolds numbers $R_e$ and at $R_e = 0$ (grey). As $R_e$ decreases, the $\dot{x}$ versus $\dot{r}$ relationship gets closer to the motility map $\dot{x} = A(r)\dot{r}$, where $A = \frac{\mu_2 m_1 - \mu_1 m_2}{(\mu_1+\mu_2)(m_1+m_2)}$ is independent of $r$.}
	\label{fig:n=0.5}
\end{figure*}

\begin{figure*}[!htbp]  
	\centering
	\subfloat[]{%
		\includegraphics[width=0.48\textwidth]{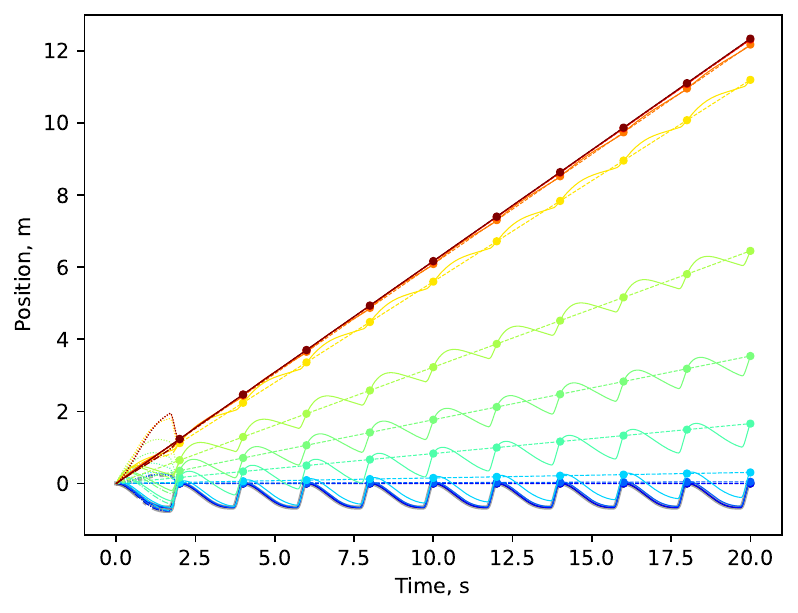}
		\label{fig:traj_n=2}}
	\hfill
	\subfloat[]{%
		\includegraphics[width=0.48\textwidth]{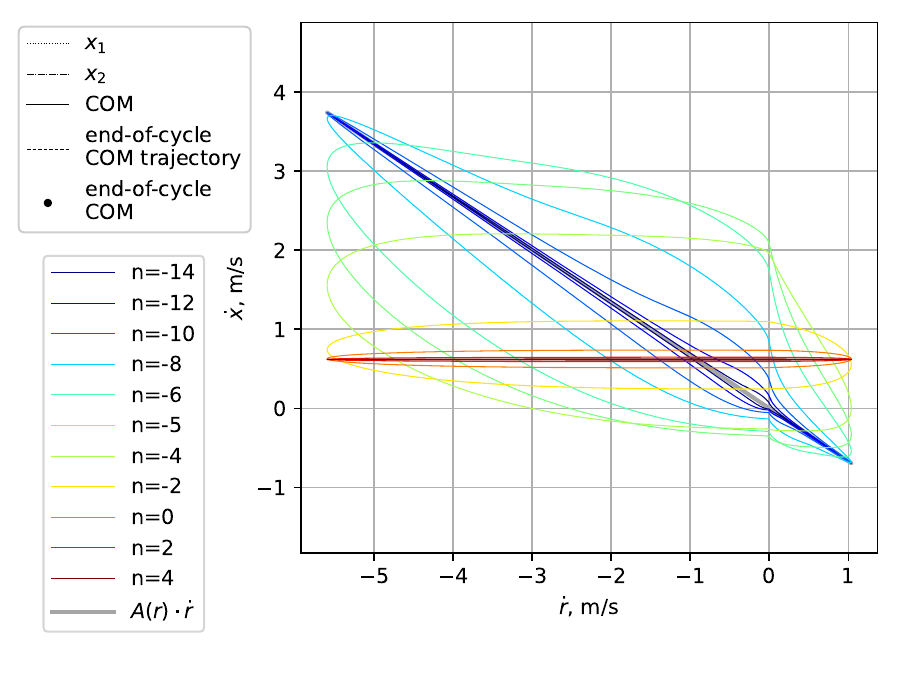}
		\label{fig:conn_n=2}}
	\caption{Inchworm dynamics with power law 2. %
	We simulated multiple Reynolds numbers and masses, holding $m_2 = 10m_1$ and $\mu_2 = \frac{\mu_1}{10}$. 
	(a) The positions of the two masses $x_1$ and $x_2$ versus time for different masses $m_1$, where $m_1 = 2^n$ kg. %
	The positions $x_1$ (dotted line), $x_2$ (dashed line), and the center of mass (COM) position (solid line) are shown. %
	The two masses oscillate causing the COM to move. %
	The net displacement over a cycle for small masses, i.e. Reynolds numbers $R_e\to 0$, goes to zero as expected. %
	When the masses are large, a net motion is produced, in the opposite direction from that for a power law with power 1/2. %
	(b) COM velocity $\dot{x}$ versus shape change velocity $\dot{r}$ at different Reynolds numbers $R_e$ and at $Re = 0$ (grey). 
	As $R_e$ decreases, the $\dot{x}$ versus $\dot{r}$ relationship gets closer to the motility map $\dot{x} = A(r)\dot{r}$, where $A = \frac{\mu_2 m_1 - \mu_1 m_2}{(\mu_1+\mu_2)(m_1+m_2)}$ is independent of $r$.}
	\label{fig:n=2}
\end{figure*}

\subsubsection{Inchworm in an \ODW\ Fluid}
Figure \ref{fig:n=1} shows the behavior of the inchworm with conventional viscous drag as mass is decreased, corresponding to decreasing effective Reynolds number $R_e$. As $R_e$ comes closer to zero, the relationship between center of mass (COM) velocity and the shape change velocity gets closer to the motility map (Figure \ref{fig:conn_n=1}), even while the within-cycle oscillations maintain the same as Figure \ref{fig:shape}.

Figure \ref{fig:n=0.5} and \ref{fig:n=2} shows the behavior of the system with a $\beta = 0.5$ and $\beta = 2$ \ODW\ viscosity (equation \ref{eqn:odw}).
These show that when $R_e$ is sufficiently large viscosity power laws away from $\beta=1$ seem to produce net motion, but $\beta>1$ and $\beta<1$ produce motion in opposite directions.
And both larger and smaller values of $\beta$ away from 1 produce more motion at sufficient $R_e$
\begin{table*}[t]
	\centering
	\resizebox{\linewidth}{!}
	{
	\begin{tabular}{lllllll}
		\hline
		Biological fluid &
		Citation &
		$\eta_0$ [Pa\,s] &
		$\eta_\infty$ [Pa\,s] &
		$\lambda$ [s] &
		$a$ &
		$n$ \\
		\hline
		Blood (tapered arteries) &
		\cite[pg 6]{mekheimer2018carreau} &
		0.056 & 0.036 & 3.313 & 2 & 0.3568 \\
		Blood (large artery) &
		\cite[Table 1]{attia2020} &
		0.161 & 0.00345 & 39.418 & 2 & 0.48 \\
		Synovial fluid (bovine/human) &
		\cite[Table 4.1, Table 4.9]{pan2014synovial} &
		3.4 &  0.02 & 5.395 & 4.775 & 0.47 \\
		Synovial fluid (Hyaluronic acid derivatives) &
		\cite[Table 2]{Chernos2017} &
		0.08 & 0.003 & 0.03 & 1.02 & 0.330 \\
		Airway mucus (lung airway mucosa) &
		\cite[pg 5, Table 1]{Bartlett2023} &
		302 & 0 & 8440 & 2 & 0.375 \\
		\hline
	\end{tabular}}
	\caption{\vspace{0.5cm}Published Carreau-Yasuda parameter sets for biological fluids.}
	\label{tab:carreau_biofluids}
\end{table*}

\subsubsection{Inchworm in a Carreau-Yasuda Fluid}
\begin{figure*}[t]  
	\centering
	\subfloat[]{%
		\includegraphics[width=0.27\textwidth]{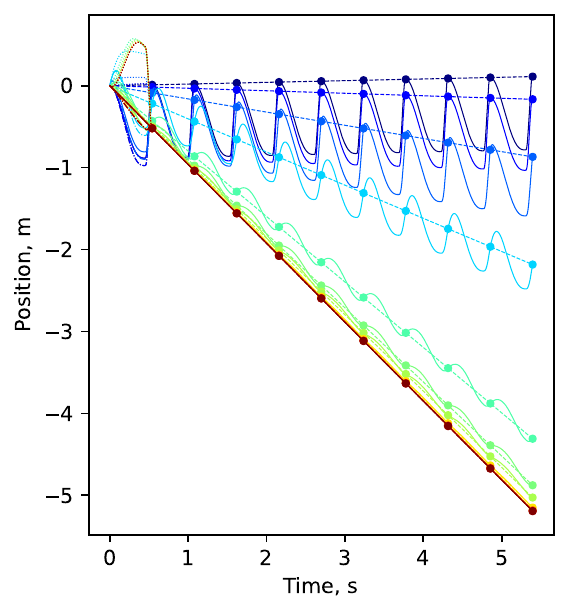}
		\label{fig:traj_n=-1_v=10}}
	\hfill
	\subfloat[]{%
		\includegraphics[width=0.27\textwidth]{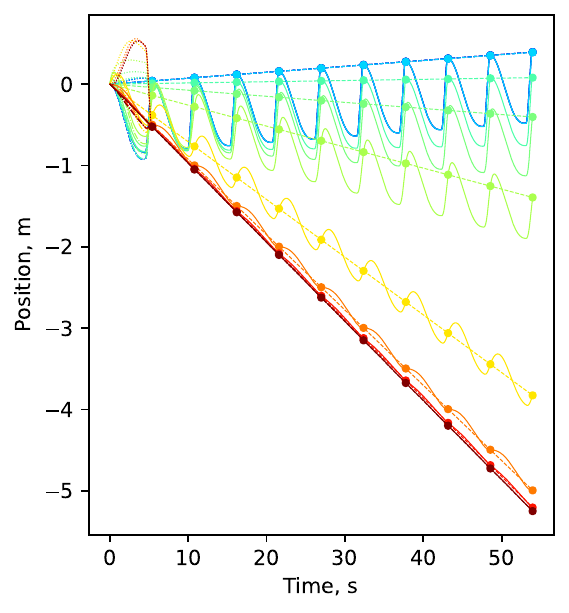}
		\label{fig:traj_n=-1_v=1}}
		\hfill
	\subfloat[]{%
		\includegraphics[width=0.4\textwidth]{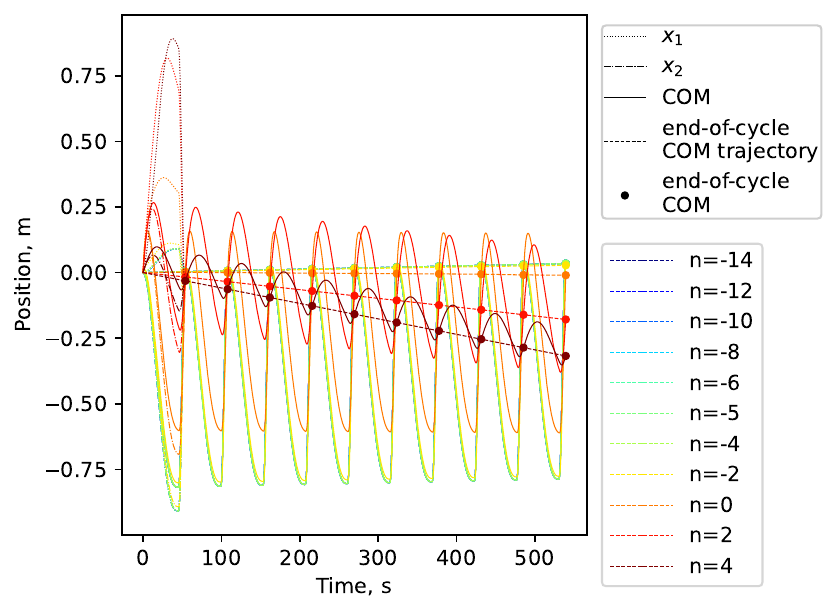}
		\label{fig:traj_n=-1_v=0.1}}
	\caption{Inchworm dynamics with Carreau-Yasuda friction with the parameters for synovial fluid (bovine/human) in \ref{tab:carreau_biofluids} at different Reynolds numbers, where $m_2 = 10m_1$, $\mu_2 = \frac{\mu_1}{10}$ and $m_1 = 2^n$ kg. %
	The positions $x_1$ (dotted line), $x_2$ (dashed line), and the center of mass (COM) position (solid line) are shown. %
	We plotted fast (period $0.1\lambda$, (a)), medium speed (period $1\lambda$, (b)), and slow (period $10\lambda$, (c)) motions.}
	\label{fig:n=-1}
\end{figure*}

Examination of equation \ref{eqn:carr} reveals that the Carreau-Yasuda fluids behave as Newtonian fluids at low and high speeds, and as \ODW\ fluids in the intermediate range.
To demonstrate the effect of Carreau-Yasuda fluid dynamics on the ability to move using reciprocal motion, we simulated inchworm using parameters of synovial fluid in Table~\ref{tab:carreau_biofluids} that takes $1\lambda, \lambda, 10\lambda$ as the total period, where $\lambda$ is the time constant in the Carreau-Yasuda model. 
We kept the time to contract at $1/5$ that of expansion.
The results can be seen in Figure~\ref{fig:n=-1}. 
In the Carreau-Yasuda fluid, even at the low mass limit corresponding to Reynolds numbers $R_e \to 0$, the inchworm can still achieve net displacement with a reciprocal gait. 
However, at the very low speeds in \ref{fig:traj_n=-1_v=0.1} and the very high speeds in \ref{fig:traj_n=-1_v=10}, the net displacement over a cycle at the smallest mass is smaller than at an intermediate speed.
We attribute this to the fact that at the low and high limits Carreau-Yasuda fluids behave more like Newtonian fluids, which would produce no net motion.

Interestingly, intermediate masses (e.g. $2^{-8}$kg) move in opposite directions at intermediate speeds \ref{fig:traj_n=-1_v=1} and at high speeds \ref{fig:traj_n=-1_v=10}, while producing comparable displacements per cycle.

\begin{figure} 
	\centering
	\subfloat[]{%
		\includegraphics[width=0.48\textwidth]{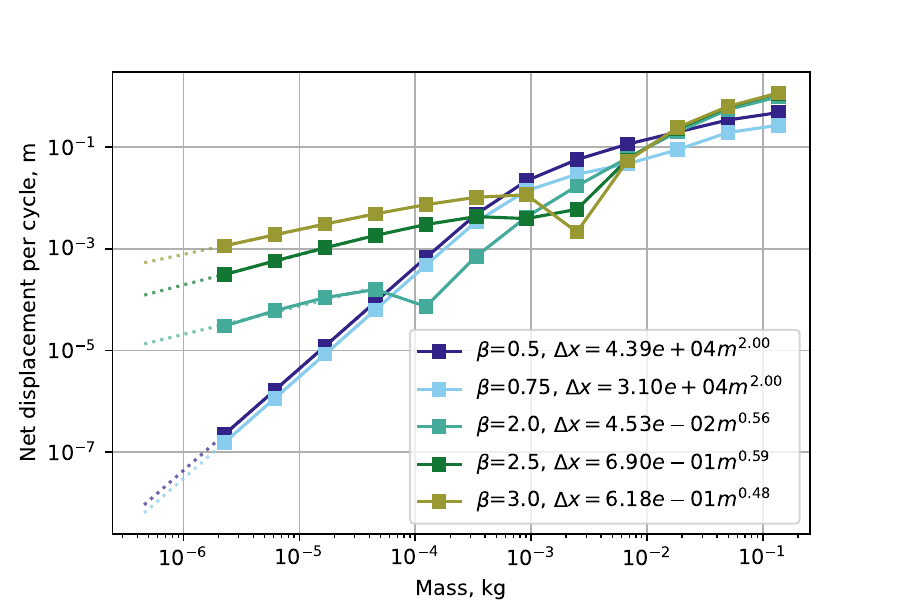}
		\label{fig:disp_power}}
	\hfill
	\subfloat[]{%
		\includegraphics[width=0.48\textwidth]{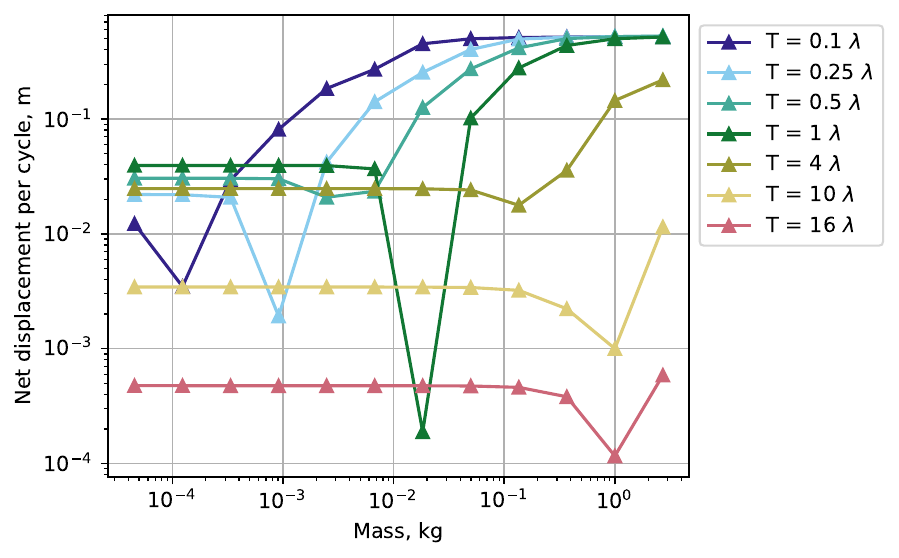}
		\label{fig:disp_carreau}}
	\caption{Inchworm COM net displacement over a cycle versus mass with different drag laws. %
	(a) We plotted results for power law models with different exponents $\beta$ (solid lines with square markers) and a linear fit to the results for small masses (dashed lines). %
	The slope at small masses indicates that displacement would go to zero when $R_e \to 0$, if the same trend continues. %
	(b) We plotted results for Carreau-Yasuda models with different period $T$ (solid lines with triangle markers). %
	These all tend to the horizontal when mass gets small, indicating a net displacement would still exist at $R_e \to 0$. }
	\label{fig:disp}
\end{figure}

Figure \ref{fig:disp} shows the absolute net displacement of the inchworm over a cycle versus mass with different power law models and the Carreau-Yasuda model. 
As $R_e$ decreases from large values, the signed displacement per cycle decreases and may cross zero at intermediate $R_e$. 
This sign reversal leads to non-monotonic behavior in the absolute displacement, producing the local minima (troughs) observed in Figure~\ref{fig:disp}.

The change of direction of motion across different $Re$ happens in both power law with $\beta > 1$ and the Carreau-Yasuda model with our selected parameters.
In the power-law case, the net displacement changes direction at higher $Re$ for larger $\beta$ values. 
In the Carreau-Yasuda case, the net displacement also changes direction at higher $Re$ for larger periods $T$.

At low $R_e$ regime, as the $Re$ decreases, the net displacement over a cycle approaches zero for all power law models, but approaches a non-zero value for the Carreau-Yasuda models. 
This is consistent with our analysis in \ref{sec:homogeneous} that the power law fluids indeed still obey the Scallop Theorem as Reynolds number $R_e\to 0$, while the Carreau-Yasuda fluids do not.
The non-zero net displacement in the Carreau-Yasuda models is the largest when the period is close to $\lambda$, and decreases as the period moves away from $\lambda$ in either direction.

%% file: discussion.tex
\section{Discussion }\label{sec:future}

Our work here showed both theoretically (in the proof in section \ref{sec:pfScal}) and by numerical experimentation (section \ref{sec:locomotion_model}) that motion in power law fluids at low Reynolds number is governed by a motility map which is homogeneous in shape change rates, and therefore these motions obey the Scallop Theorem.
Our inchworm model allowed us to see how the low Reynolds number limits appear when inertias get smaller, in particular showing that with non-vanishing Reynolds numbers, motions in \ODW\ fluids occur in opposite directions for exponents larger than one and exponents smaller than one.
We also motivated the appearance of motility maps in Coulomb friction systems as a limiting case of the motility maps of power-law friction motility maps.
Finally, we explored the breakdown of the Scallop Theorem in Carreau-Yasuda fluids, showing that in those fluids reciprocal motion can produce net displacement even at Reynolds numbers approaching zero.
Surprisingly, for intermediate Reynolds numbers our Carreau-Yasuda simulations suggest the direction of motion can be controlled by the period of oscillation.

\subsection{$\Hf^1$ Motility Maps extend the reach of Geometric Mechanics}

Our discovery of motility maps for non-linear friction motivates extending the many tools of geometric mechanics developed to understand biological swimming and robot locomotion \cite{hatton2013, hatton2013-2, rieser2024} in terms of linear motility maps to $\Hf^1$ motility maps.
The results could significantly improve our ability to analyze and design locmotion systems in non-Newtonian fluids.
Since our results apply to many fluids of medical importance, future applications in medical microswimmers seem to be a natural choice.

\subsection{Motility Maps for Coulomb Friction Systems?}

The puzzling efficacy of motility maps for Coulomb friction systems such as multi-legged animals and robots has been noted before  \cite{Chong2022, zhao2022,  wu2025}. 
Our results here suggest for the first time a path to mathematical proofs for understanding this phenomenon as the $\gamma\to 0$ limit of $\Hf^\gamma$ friction laws.
The steps of this proof would require understanding when the Lagrangian models of power-law friction admit motility map solutions; understanding when these solutions are convergent at the $\gamma\to 0$ limit; and understanding how these limits express themselves upon the structure of the motility maps -- are the limiting motility maps truly linear? are they $\Hf^1$? are there other structural properties we could derive, e.g. the Finsler structures used in \cite{hatton2025}.

\subsection{Locomotion in Carreau-Yasuda fluids}

Using a one degree of freedom inchworm model, we demonstrated that Carreau-Yasuda viscosity can enable systems with only one degree of freedom to generate net motion in a low Reynolds number environment, and to reverse their direction of motion by changing speeds in intermediate Reynolds numbers.
This result extends to systems with any number of degrees of freedom and suggests that some nonlinear viscosity laws that are not power laws can allow the ``Purcell Scallop Theorem'' to be violated. 
Our finding suggests that micro-swimmers designed for medically important Carreau-Yasuda fluids, such as those in Table~\ref{tab:carreau_biofluids} could use 1-DoF actuators to produce a net motion -- an important simplification for systems where every DoF is a challenge.
One may envision a microswimmer with two uncoordinated, remotely actuated DoF -- one used to steer and one used to produce thrust, where each DoF produces motion either forward or back based on its driving frequency.
According to conventional Scallop Theorem informed thinking such a microswimmer would be unable to move as it can only perform reciprocal motions.
Our results suggest otherwise, and open the gate to new families of designs.

\subsection{Summary}

Our work here extended the reach of motility maps to power-law friction, whether in the form of \ODW\ fluids or as relaxations of Coulomb friction.
We have shown that Purcell's famous Scallop Theorem extends its reach to these systems as well.
However, inertial effects and Carreau-Yasuda viscosity can each take a system making reciprocal motions outside the reach of the Scallop Theorem and produce net displacement. 
The motility maps that arise from power laws have less structure than those arising from linear viscosity -- they are merely homogeneous instead of being fully linear.
We hope to join the community in exploring this class of models in future work.